\documentclass[letterpaper, 10 pt, journal, twoside]{IEEEtran}

\usepackage{decar-common}
\usepackage{decar-dynamics}
\usepackage{decar-lie}
\usepackage{decar-post}
\usepackage{tabularx}
\usepackage{booktabs}
\usepackage{subcaption}
\usepackage[normalem]{ulem}

\DeclarePairedDelimiterXPP\Aver[1]{\mathbb{E}}{[}{]}{}{

#1
}

\newsavebox{\subfigbox}
\DeclareMathOperator{\tr}{tr}
\makeatletter
\AtBeginDocument{
  \check@mathfonts
}
\makeatother

\begin{document}


\fontdimen16\textfont2=\fontdimen17\textfont2
\fontdimen13\textfont2=5pt

\captionsetup{font=footnotesize}
\captionsetup[sub]{font=footnotesize}
\title{Bayesian Filtering for Homography Estimation}

\author{Arturo~Del~Castillo~Bernal,
Philippe~Decoste,
        and~James~Richard~Forbes%
        
\thanks{This paper was recommended for publication by Editor 
P. Vasseur upon evaluation of the Associate Editor and Reviewers' comments. 
    This work was supported by the NSERC Discovery Grant program, and the Canadian Innovation Fund (CFI) program.
This work was also supported in part by funding from the Innovation
for Defence Excellence and Security (IDEaS) program from the Department
of National Defence (DND). 
Any opinions and conclusions in this work
are strictly those of the authors and do not reflect the views, positions,
or policies of - and are not endorsed by - IDEaS, DND, or the Government
of Canada.
}%

\thanks{A. Del Castillo Bernal, P. Decoste, and J. R. Forbes are with the Department of
Mechanical Engineering, McGill University, Montreal, QC, Canada, H3A~0C3.
{\tt \footnotesize arturo.delcastillo@mail.mcgill.ca}}%

\thanks{Manuscript received June 14, 2021; Revised September 25, 2023; Accepted October 13, 2023.}
\thanks{Digital Object Identifier (DOI): see top of this page.}
}

\markboth{IEEE Robotics and Automation Letters. Preprint Version. Accepted October, 2023}
{Del Castillo \MakeLowercase{\textit{et al.}}: Bayesian Filtering for Homography Estimation}

\maketitle

\begin{abstract}
    This paper considers homography estimation in a Bayesian filtering framework using rate gyro and camera measurements. 
    The use of rate gyro measurements
    facilitates a more reliable estimate of homography in the presence of occlusions, 
    while a Bayesian filtering 
    approach generates both a homography estimate along with an uncertainty. 
    Uncertainty information opens the door to adaptive filtering approaches,
    post-processing procedures, and safety protocols. In particular, 
    herein an iterative extended Kalman filter and an interacting multiple model (IMM)
     filter are tested using both simulated and experimental datasets.
     The IMM is shown to have good consistency
    properties and better \textcolor{black}{overall} performance when compared to the
     state-of-the-art homography nonlinear deterministic observer in both simulations \textcolor{black}{
    and experiments.}
\end{abstract}


\section{Introduction and Related Work}
\label{sec:intro}

\IEEEPARstart{A} homography is a mapping that relates two views of the same planar scene. It is exploited in robotics applications when the
structure of the environment is sufficiently planar, such as indoor hallways, manmade experimental settings, and aerial coverage. 
Robotics applications that have successfully used homography include
  visual-servoing \cite{benhimane2007homography}, image stabilization 
  \cite{hua2019feature}, 
ego-motion \cite{grabe2015nonlinear,hua2018attitude}, and monocular SLAM initialization  \cite{mur2015orb}.
Homographies can better explain the structure of planar scenes and low parallax compared to
 using 
the epipolar constraint \cite{hartley2003multiple}, which is commonly employed
for unstructured scenes.

Homography is usually estimated using feature correspondences between
a pair of images, such as points, lines, conics or a combination thereof \cite{agarwal2005survey,kaminski2004multiple}. 
Direct methods \cite{benhimane2004real}
and learning-based approaches  \cite{detone2016deep,nguyen2018unsupervised} 
are alternative means to estimate homography. 
All of these methods consider 
camera measurements independently and ignore any temporal correlations, thus making them susceptible
to failures in the presence of occlusions or lack of feature correspondences \cite{hartley2003multiple}. 

In \cite{6160724},
temporal information is exploited to estimate homography by designing a nonlinear
deterministic observer. 
\textcolor{black}{This observer is able to operate even during
occlusions by propagating prior estimates using angular velocity measurements,
thus providing
a more robust estimate of the homography}.
 This comes at the cost of assuming the camera's velocity is constant
  parallel to the plane or exponentially converging towards to the plane. 
 Further, homography is parameterized as an element of the special linear group $SL(3)$. 
Lie group properties are exploited in the observer structure
to prove local asymptotic stability under various assumptions.

Nonlinear deterministic observers aim to show
\textit{a priori} stability properties, and they 
do not take into account stochastic processes, such as noise
in sensor measurements, which is present in all real systems. 
This paper leverages the tools of Bayesian filtering 
 to take into account noise statistics to produce 
an accurate and consistent estimate with a covariance describing 
 the error distribution. 
A Bayesian filtering approach opens the door for procedures such as smoothing \cite[Ch. 8]{Saerkkae2013},
 loop-closure detection \cite{caballero2007homography}, 
 adaptive approaches \cite[Ch. 11]{bar2001estimation}, or simply 
 monitoring the filter's quality \cite[Sec. 5.4]{bar2001estimation}.

 \begin{figure}[t!]
    \centering
    \includegraphics[width = 0.9 \linewidth]{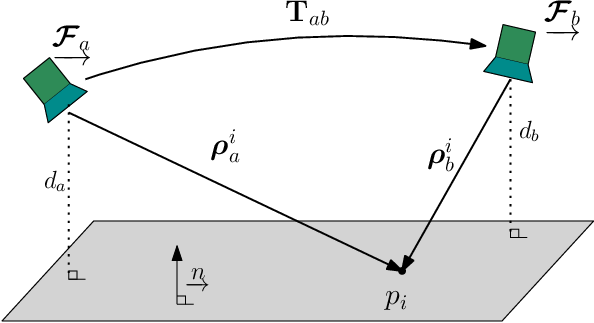}
    \caption{The transformation of the observed position of points on a 
    plane by a camera viewing the scene from 
    different perspectives is described by a homography matrix.}
    \label{fig: homography}
 \end{figure}

This paper presents a Bayesian filtering approach to homography estimation
 using rate gyro and camera measurements \textcolor{black}{in a tightly-coupled manner, with covariance information.}
  The same assumption on the camera's 
 inertial-frame velocity from \cite{6160724} is made to enable the 
 use of a \textcolor{black}{rate gyro and a specific motion model}. Because this assumption is violated from time to
   time in practice, focus is placed on a specific type of Bayes filter,
    the interacting multiple model (IMM) filter. 
The IMM filter adapts the noise level of the process model when 
the velocity assumption of the process model is violated. This approach
is similar to the IMM application found in \cite{li1994recursive}. The IMM 
used here is composed
of two iterated extended Kalman filters (EKFs).
This paper clearly demonstrates improved performance,
 along with consistency, in simulation, as well as
  better overall performance in experiments
 of the IMM relative to a nonlinear deterministic observer \cite{hua2019feature}. 
 As such, this paper's contribution is the combination of 
 1) a Bayesian filtering framework, 2) utilizing the camera
  velocity assumption of \cite{6160724} in a probabilistic setting, thus providing a simple 
  process model and enabling the use of the rate gyro, 3) considering the 
  IMM structure to account for violations of the camera
   velocity assumption, and 4) simulating and experimentally testing the
  proposed IMM filter relative to a nonlinear deterministic observer.

Deterministic and Bayesian approaches to homography 
estimation are not mutually exclusive nor competitors. The choice 
to use one or the other, or perhaps both in ``primary" and ``back-up"
 roles, or even both in synergy, will be application dependent. 
 In situations where a computationally simple observer is needed,
  but covariance information is not needed, a deterministic observer 
  is a natural choice. On the other hand, in situations where covariance
  information is needed, and computational resources are not so limited,
a Bayesian filter is appropriate. As such, this paper does
not advocate for the displacement of \cite{6160724,hua2018attitude,hua2019feature}
 and similar work. 
 Rather, this paper builds on \cite{6160724} by providing a means
  to use both a rate gyro and camera to generate a homography
   estimate along with a covariance in a way that accounts
    for the limitations of the assumed process model of \cite{6160724}.


\section{Preliminaries}
    Herein, $\mbf{x} \sim \mathcal{N}(\mbs{\mu}, \mbs{\Sigma})$ is used 
    to denote a Gaussian random variable $\mbf{x}\in\rnums^n$ with mean $\mbs{\mu}\in\rnums^n$ and 
    positive definite covariance
    matrix $\mbs{\Sigma}\in\rnums^{n\times n}$. 
    The notation $\check{(\cdot)}$ and $\hat{(\cdot)}$ 
    denotes the predicted and measurement-corrected
    state estimates, respectively.  The superscripts  $(\cdot)^i$  are indices,
    not exponents. For a random variable $\mbf{X}$
    evolving in a matrix Lie group $G$,
    a ``Gaussian'' distribution is represented by 
    $
        \mbf{X} = \exp(\mbs{-\xi}^\wedge)\mbfbar{X}, 
    $
     where $\mbs{\xi}\sim \mathcal{N}(\mbf{0}, \mbs{\Sigma})$, and $\mbfbar{X}$ is the mean~\cite{barfoot2014associating}. 

    For a generic function $\mbf{g}: G \to \mathbb{R}^m$ of a Lie group element, the shorthand notation for a Lie derivative is defined as 
\begin{equation}
\left.\frac{D \mbf{g}(\mbf{X})}{ D \mbf{X}} \right|_{\mbfbar{X}}
 \triangleq \left.\frac{\p \mbf{g}( \exp(-\mbs{\xi}^\wedge)\mbfbar{X})}{\p \mbs{\xi}} \right|_{\mbs{\xi} = \mbf{0}}.
\end{equation}

    \subsection{Homography}
    \label{sec:homography}

    As shown in Figure~\ref{fig: homography}, a homography mapping relates features on a plane from two distinct views.
    A point $p_i$ lying on a plane has coordinates resolved in the camera frame $\mathcal{F}_{a}$ 
    given by $\mbs{\rho}_a^i = \bbm x & y & z \ebm^\trans$. The normalized image coordinates are 
    \begin{equation}
        \mbf{p}_a^i = \bbm x/z \\ y/z \\ 1   \ebm = \f{1}{z}\mbs{\rho}_a^i.
    \end{equation}
    It is possible to obtain 
    the projection of $\mbf{p}_a^i$ on the image plane $\mbf{q}_a^{i}$
    by using the intrinsic parameter matrix $\mbf{K}$, written as
    \begin{align}
        \mbf{q}_a^i &= \bbm u \\ v \\ 1   \ebm = \underbrace{\bbm f_u & 0 & c_u 
        \\ 0 & f_v & c_v \\ 0 & 0 &1 \ebm}_\text{$\mbf{K}$} \mbf{p}_a^i,
    \end{align}
    where $f_u$ and $f_v$ are the horizontal and vertical focal lengths 
    and $c_u$ and $c_v$ are the optical center coordinates in pixels.
    As the camera's pose changes by $\mbf{T}_{ab} \in SE(3)$ \cite[Sec. 8.1]{barfoot2017state},
    it is possible to relate a new observation $\mbf{q}_b^i$ 
    of $p_i$ to $\mbf{q}_a^i$ by a \textit{homography matrix} $\mbf{H}_{ab}$ \cite{6160724},
    \begin{align}
        \hspace{-5pt}\mbf{q}_b^i &= \f{z_a}{z_b}\mbf{K}\left(\mbf{C}_{ab} - 
        \f{1}{d_b}\mbf{r}_{a}^{ba} \mbf{n}_b^\trans \right)\inv\mbf{K}\inv \mbf{q}_a^i 
        =\mbf{K}\mbf{H}_{ab}\inv \mbf{K}\inv \mbf{q}_a^i, \label{eq: point_corr}
    \end{align}
    where $\mbf{C}_{ab} \in SO(3)$ is the direction cosine matrix
     that encodes the orientation of $\mathcal{F}_a$ 
    relative to $\mathcal{F}_b$, $\mbf{r}_a^{ba}$ is the position of the origin of $\mathcal{F}_b$
     relative
     to the origin of $\mathcal{F}_a$ resolved in $\mathcal{F}_a$, $\mbf{n}_b$ is the normalized plane direction
     resolved in $\mathcal{F}_b$, and $d_b$ is the orthogonal distance 
     from the camera position to the plane.
    The definition $\gamma \triangleq z_a / z_b$ will be used throughout the paper.

    \subsection{Special Linear Group}
    \label{sec:sl3}

    There are 6 degrees of freedom associated with rotation and translation, 
    and 3 degrees of freedom associated with relative distance. 
    However, a homography matrix only has 8 degrees of freedom, not 9,
    because it is defined only up to the scalar $\gamma$. One way
    to parameterize homography is by the Special Linear Group $SL(3)$ \cite[Chap. 1]{hall2013lie},
    \begin{equation}
        SL(3) \equiv  \{\mbf{H}\in \rnums^{3\times3}\, \vert \,\det \mbf{H} = 1  \}.
    \end{equation}
    Any nonsingular matrix $\mbf{X}\in \rnums^{3\times 3}$ can be projected
    onto $SL(3)$ by
    $
        \mbf{X}/ \left(\det \mbf{X}\right)^{1/3}\in SL(3).
    $
    As with any Lie group, there is a Lie algebra associated with $SL(3)$, defined as
    \begin{equation}
        \mathfrak{sl}(3) \equiv  \{\mbs{\Xi}\in \rnums^{3\times3}\, \vert \, \tr( \mbs{\Xi}) = 0  \},
    \end{equation}
    \textcolor{black}{where $\tr(\cdot): \mathbb{R}^{3 \times 3} \to \mathbb{R}$ is the trace operator.}
    The matrix exponential is a surjective mapping that relates the Lie algebra to the Lie group,
        $\mbf{X} = \exp\left(\mbs{\Xi}\right)$. For $SL(3)$, the inverse mapping
        is defined only for elements near identity,
        $\mbs{\Xi} = \log\left(\mbf{X}\right)$.
    A vector can be uniquely associated to an element of $\mathfrak{sl}(3)$ by
    $(\cdot)^\wedge: \mathbb{R}^8 \to \mathfrak{sl}(3)$, which is defined as
     \cite{eade2014lie},
    \begin{equation}
        \mbs{\xi}^\wedge
        = \bbm \xi_4+\xi_5 & -\xi_3+\xi_6 & \xi_1 \\
            \xi_3+\xi_6 & \xi_4-\xi_5 & \xi_2 \\
            \xi_7 & \xi_8 & -2\xi_4   \ebm.
    \end{equation}
    The operator $(\cdot)^\vee: \mathfrak{sl}(3) \to \mathbb{R}^8$ is defined as the inverse
    of $(\cdot)^\wedge$.
    The adjoint operator $\mathrm{Ad}: SL(3) \times \mathfrak{sl}(3) \rightarrow \mathfrak{sl}(3) $ is defined as 
    $
        \mathrm{Ad}(\mbf{H})\mbs{\Xi} = \mbf{H}\mbs{\Xi}  \mbf{H}\inv,
    $
    where $\mbs{\Xi} \in \mathfrak{sl}(3) $. The adjoint matrix $\mbf{Ad}(\mbf{H})$ can be obtained by
        $\left(\mathrm{Ad}(\mbf{H})\mbs{\Xi}\right)^\vee = \mbf{Ad}(\mbf{H})\mbs{\xi}$.
    The Lie bracket $[\cdot, \cdot]: \mathfrak{sl}(3)\times\mathfrak{sl}(3) \to \mathfrak{sl}(3)$ is defined as
   $
        \left[\mbs{\Xi}_1, \mbs{\Xi}_2 \right] = \mbs{\Xi}_1\mbs{\Xi}_2 - \mbs{\Xi}_2\mbs{\Xi}_1,
  $
    and a \textit{little adjoint} matrix can be defined also as,
        $\left[\mbs{\Xi}_1, \mbs{\Xi}_2 \right]^\vee = \mbf{ad}(\mbs{\xi}_1)\mbs{\xi}_2$.
    The $(\cdot)^\odot: \mathbb{R}^3 \to \mathbb{R}^{3 \times 8}$ operator
    is defined as
        $\mbs{\xi}^\wedge\mbf{p} = \mbf{p}^\odot \mbs{\xi}$,
    where $\mbf{p} \in \rnums^{3}$.

    \subsection{Homography Kinematics}
    \label{sec: kinematics}
    The kinematics of $\mbf{H}_{ab}$ are \cite{6160724}
    \begin{align}
        \mbfdot{H}_{ab}&= \mbf{H}_{ab} \left({\mbs{\omega}_b^{ba}}^{\times} -\f{\mbf{v}_b^{ba} \mbf{n}_b^\trans}{d_b}
        +\f{\mbf{n}_b^\trans \mbf{v}_b^{ba} }{3 d_b} \eye \right), \label{eq: H_kinematics}
    \end{align}
    where ${\mbs{\omega}_b^{ba}}$ is the angular velocity of the body with respect to (\wrt) the inertial frame
    resolved in $\mathcal{F}_b$, $\mbf{v}_b^{ba}$ is the linear velocity of the body \wrt the inertial frame
    resolved in $\mathcal{F}_b$, and 
     $(\cdot)^\cross : \rnums^3 \rightarrow \mathfrak{so}(3)$,
     where $ \mathfrak{so}(3)$ is the Lie algebra
    of $SO(3)\subset SL(3)$. The kinematic model~\eqref{eq: H_kinematics}
     depends on the plane parameters $\mbf{n}_b$ and $d_b$,
    which are unmeasurable. However, by assuming that 
    $\mbf{s}_{a} \triangleq  \f{\mbfdot{r}_{a}^{ba}}{d_b} $ is constant in time as in
    \cite{6160724,hua2019feature},
    a simplified kinematic model can be derived. Setting
     $\mbs{\Gamma}_{ab} = -\f{\mbf{v}_b^{ba} \mbf{n}_b^\trans}{d_b}
    +\f{\mbf{n}_b^\trans \mbf{v}_b^{ba} }{3 d_b} \eye \in \mathfrak{sl}(3)$, 
    where $\mbf{1}$ is the identity matrix,
    the assumption that $\mbf{s}_{a}$ 
    is constant yields the simplified process model
    \begin{align}
        \dot{\mbf{H} }_{ab}&= \mbf{H}_{ab}\left( {\mbs{\omega}_b^{ba}}^{\cross} + \mbs{\Gamma}_{ab}  \right), \qquad 
        \dot{\mbs{\Gamma}}_{ab} = \left[\mbs{\Gamma}_{ab}, \,\, {\mbs{\omega}_b^{ba}}^{\cross}\right]. \label{eq:process_model}
    \end{align}
    This simplified process
    model depends only on angular velocity, which is easily measured 
    by a rate gyro. However, assuming $\mbf{s}_a$ is constant
     restricts the trajectory
      of the camera to have constant velocity parallel
    to the plane, or to  
    exponentially converge towards the plane \cite{6160724}.

    \subsection{Interacting Multiple Model}
    \label{sec: imm}
    
    An IMM manages multiple models 
    of a dynamic system in estimation tasks. The model at time step $k$
    is assumed to be among $n$ possible models $\theta(k) = \{\theta_i\}_{i=1}^{n}$.
    The model switching in the system is assumed
    to be a Markov chain with known transition probabilities 
    \begin{equation}
        p_{ij} \triangleq P\left(\theta(k)=\theta_j \vert \theta(k-1) =\theta_i\right).
    \end{equation}
    In robotics applications,
    the IMM is used to mix the estimates of
    multiple filters each with their own process model of the system, 
    where the $i^\textrm{th}$ process model $\mbf{f}_i(\cdot)$ is written
    \begin{align}
        \mbf{x}_{k}^i &= \mbf{f}_i(\mbf{x}_{k-1}^i, \mbf{u}_{k-1}).
    \end{align}
    To avoid the exponential complexity of accounting for all possible
    $n^k$ combination of models at time step $k$, the IMM runs $n$ process models
     in parallel 
    at all times, 
    each with a weight $w^i_k \triangleq P(\theta(k)= \theta_i \vert \mbf{y}_{1:k})$.
    When a new measurement $\mbf{y}_k$ arrives, 
    the IMM executes the following three steps \cite[Sec. 11.6]{bar2001estimation}. 
        \subsubsection{Interaction}
        \textit{Mixing probabilities} are calculated by
        \begin{align}
            \begin{split}
                \mu^{ij}_{k-1}&\propto 
                P\left( \theta(k)=\theta_j \vert \theta(k-1) =\theta_i\right)\times  \\
                &\quad\quad P\left( \theta(k-1)=\theta_i \vert \mbf{y}_{1:k-1}\right), \\
                \mu^{ij}_{k-1}&=\f{1}{c}p_{ij}w^i_{k-1},
                \end{split}
        \end{align}
        where $c$ is a normalization constant. 
        The mixing probabilities represent how likely 
        a switch is to happen given the history
        of measurements or current knowledge of the trajectory.
        Every model $i$ computes mixing probabilities
        with the rest of the models including itself.
        \subsubsection{Mixing} Since 
        it is assumed
        that the states are Gaussian distributed, a Gaussian mixture
         procedure, considering the mixing probabilities,
         is carried out to update the state and covariance estimate of each filter,
        \begin{align}
            \mbfcheck{x}_{k-1}^{i} &\leftarrow \sum_j^n \mu^{i j}_{k-1}\mbfcheck{x}_{k-1}^j, \label{eq: gm_mean}\\
            \begin{split}
            \mbfcheck{P}_{k-1}^{i} &\leftarrow \sum_j^n \mu^{i j}_{k-1}(
                \mbfcheck{P}_k^j +
                 (\mbfcheck{x}_{k-1}^{j}-\mbfcheck{x}_{k-1}^{i})(\mbfcheck{x}_{k-1}^{j}-\mbfcheck{x}_{k-1}^{i})^\trans ).
             \label{eq: gaussian_mixture_cov}
            \end{split}
        \end{align}
        The mixing is straightforward for states in vector space. For states that evolve 
        in a Lie group, a reparametrization
        of each estimate's  $\mbfcheck{X}_{k-1}^j$ distribution is needed 
        about
        a reference point, in this case $\mbfcheck{X}_{k-1}^i$, to perform a Gaussian mixture 
        \cite{cesic2017mixture}. To this end,
        \begin{align}
            \begin{split}
            \mbs{\epsilon}_{k-1}^{ji} &=   \log\left(\mbfcheck{X}^i_{k-1} {\mbf{X}^j}\inv_{k-1}\right)^\vee \\
            &\approx   \mbsbar{\epsilon}_{k-1}^{ji} + \mbf{J}^{r}(\mbsbar{\epsilon}_{k-1}^{ji})\inv 
            \mbs{\xi}_{k-1}^j,
            \end{split}
            \label{eq: bch} \\
            \mbfcheck{P}_{k-1}^{ji} &=  \mbf{J}^{r}(\mbsbar{\epsilon}_{k-1}^{ji})\inv
            \mbfcheck{P}_{k-1}^{j} \mbf{J}^{r}(\mbsbar{\epsilon}_{k-1}^{ji})^{-\trans},
        \end{align}
        where $\mbsbar{\epsilon}_{k-1}^{ji}$ and $\mbfcheck{P}_{k-1}^{ij}$ are the reprojected mean
        and covariance of $\mbfcheck{X}_{k}^j$ about $\mbfcheck{X}_{k-1}^i$. Equation~\eqref{eq: bch} 
        is a first-order approximation,
         with $\mbf{J}^{r}(\mbsbar{\epsilon}_{k-1}^{ji})$ 
        being the right group Jacobian of $SL(3)$ about $\mbsbar{\epsilon}_{k-1}^{ji}$, calculated using a backward finite-difference.
        By applying~\eqref{eq: gaussian_mixture_cov} and~\eqref{eq: gm_mean}, 
        the reprojected means and covariances $\mbsbar{\epsilon}_{k-1}^{ji}$ and $\mbfcheck{P}_{k-1}^{ij}$ 
         can be combined to produce $\mbsbar{\epsilon}_{k-1}^{i}$ 
        and $\mbfcheck{P}_{k-1}^{i}$. To finalize the Gaussian mixture on $SL(3)$,
        these values are then projected onto the $SL(3)$
         manifold as follows,
         \begin{align}
            \mbfcheck{X}_{k-1}^i &\leftarrow \exp\left(- \mbsbar{\epsilon}_{k-1}^{i}\right) \mbfcheck{X}_{k-1}^i, \\
            \mbfcheck{P}_{k-1}^i &\leftarrow 
            \mbf{J}^{r}(\mbsbar{\epsilon}_{k-1}^{i})\mbfcheck{P}_{k-1}^{i}
            \mbf{J}^{r}(\mbsbar{\epsilon}_{k-1}^{i})^\trans.
        \end{align}
        \subsubsection{Correction} Once each filter has an updated mixed state estimate, the likelihood
        that filter $i$ generated
        the measurement is computed to update the weights of the filters,
        \begin{align}
            \hspace{-10pt}\Lambda^i_k = p(\mbf{y}_k \vert \mbf{X}_{k-1}^i, \theta_{k-1}), \quad
            w^i_k = \f{1}{c}\Lambda^i_k \sum_j^n w^{j}_{k-1} p_{ij},
        \end{align}
    where $c$ is a normalization constant.
    A more in-depth description of the IMM can be found in \cite[Sec. 11.6]{bar2001estimation}.

\section{Homography Estimation using the IMM Filter}
\label{sec:methodology}

The states to be estimated are $\mbf{H}_{ab}$ and $\mbs{\Gamma}_{ab}$. An IMM is implemented employing two (iterated)
 EKFs. Both EKFs share the same measurement and process model,
but have different process model noise levels, similar to \cite{li1994recursive}. Although the true
 noise levels are unknown in this work, 
the same principle will be used to manage how 
confident the filter is in the assumption that $\mbf{s}_{a}$ is constant.
To implement an EKF, the Jacobians of the process and measurement 
models must be obtained. 
 \subsection{Process Model}
The process model has access to a rate gyro measurement $\mbf{u} = \mbs{\omega}^{ba}_b + \mbf{w}$ resolved in
the body frame $\mathcal{F}_b$. It is assumed that the rate gyro is unbiased and
 is only corrupted by white Gaussian noise $\mbf{w}(t)  \sim \mathcal{N}\left(\mbf{0}, \mbc{Q}_g\delta(t-\tau)\right)$.  
From \eqref{eq:process_model},
 the process model is
\begin{align}
    \dot{\mbf{H} }_{ab}&= \mbf{H}_{ab}\left( (\mbf{u}^\cross -\mbf{w}^{\cross}) + \mbs{\Gamma}_{ab}    \right), \\
    \dot{\mbs{\Gamma}}_{ab} &= \left[\mbs{\Gamma}_{ab}, \mbf{u}^\cross -\mbf{w}^{\cross}\right]+{\mbf{w}^m}^\wedge. \label{eq: G_kin_noise}
\end{align}
As previously mentioned, assuming $\mbf{s}_{a} $ is 
constant in time
 comes at the
cost of process model inaccuracies when the assumption is broken. One way to account for
 modelling errors is adding a noise term $\mbf{w}^m(t) \sim
 \mathcal{N}\left(\mbf{0}, \mbc{Q}_m\delta(t-\tau)\right)  \in \rnums^8$.
The power spectral density (PSD) is modelled as
$\mbc{Q}_m = \sigma_m^2 \eye$. The size of $\sigma_m^2 $ 
represents how confident the filter is on the motion model for $\mbs{\Gamma}_{ab}$. 

The IMM introduced in Section~\ref{sec: imm} is a common adaptive method that weights
the available motion hypotheses on a Bayesian framework to produce an estimate. 
The idea is to have two similar models, with different values of $\sigma_m^2 $. 
Low values accommodate for scenarios where the assumption 
that $\mbf{s}_a$ is constant is respected, and higher values can deal with
scenarios where the motion violates this assumption.
The IMM
should provide an estimate with better consistency properties than using a single filter in cases where
the trajectory is more varied, switching from slow to aggressive maneuvers.

To linearize the process model, a right-invariant error definition is used, where 
\begin{align}
    \exp(\delta \mbs{\xi}^\wedge) = \mbfbar{H}\mbf{H}\inv , \label{eq: homography_error} \qquad
    \delta\mbs{\gamma}^\wedge = \mbs{\Gamma}-\mbsbar{\Gamma}, 
\end{align}
applying a Taylor series expansion, the linearized process model evolving in the Lie algebra is
\begin{subequations}
    \label{eq: linear_lie}
    \begin{align}
        \delta\dot{\mbs{\xi}}^\wedge_{ab}&= -\mathrm{Ad}\left( \mbfbar{H}_{ab}  \right)\delta\mbs{\gamma}^\wedge_{ab} +\mathrm{Ad}\left( \mbfbar{H}_{ab}  \right)\delta\mbf{w}^\cross , \\
        \delta\dot{\mbs{\gamma}}_{ab}^\wedge &=  -\mathrm{ad}\left( \mbf{u} \right)\delta\mbs{\gamma}_{ab}^\wedge  - \mathrm{ad}\left(\mbsbar{\gamma}_{ab}\right)\delta\mbf{w}^\cross+\delta{\mbf{w}^m}^\wedge .
    \end{align}
   \end{subequations}
When applying the linear $(\cdot)^\vee$ operator on 
\eqref{eq: linear_lie}, $\delta\mbf{w}^\cross $ has to be addressed. 
A projection matrix $\mbf{B}$ can be found such that, ${\delta\mbf{w}^\cross}^\vee =
 \mbf{B}\delta\mbf{w} $. As such, the linearized process model is thus 
 \begin{subequations}
    \label{eq: linear}
\begin{align}
    \delta\dot{\mbs{\xi}}_{ab}&= -\mathbf{Ad}\left( \mbfbar{H}_{ab} 
     \right)\delta\mbs{\gamma}_{ab} +\mathbf{Ad}\left( \mbfbar{H}_{ab} 
      \right)\mbf{B}\delta \mbf{w} , \\
    \delta\dot{\mbs{\gamma}}_{ab}&=  -\mathbf{ad}\left( \mbf{u} 
    \right)\delta\mbs{\gamma}_{ab} 
     -\mathbf{ad}\left(\mbsbar{\gamma}_{ab}\right)\mbf{B}\delta \mbf{w}+\delta\mbf{w}^m .
\end{align}
\end{subequations}
Setting $\delta\mbf{x}_{ab} = \bbm \delta{\mbs{\xi}_{ab}}^\trans  &   \delta{\mbs{\gamma}_{ab}}^\trans \ebm ^\trans$, 
and $\delta\mbf{w}_{ab} = \bbm \delta\mbf{w} ^\trans  &   \delta{\mbf{w}^m}^\trans \ebm ^\trans$, the linearized process model can
be written as,
\begin{align}
    \delta\dot{\mbf{x}}_{ab} &=  \underbrace{\bbm \mbf{0} & -\mathbf{Ad}\left( \mbfbar{H}_{ab}  \right) \\
      \mbf{0} &  -\mathbf{ad}\left( \mbf{u} \right) \ebm}_\textrm{$\mbf{A}$} \delta\mbf{x}_{ab}  +
      \underbrace{\bbm  \mathbf{Ad}\left( \mbfbar{H}_{ab}  \right)\mbf{B} & \mbf{0} \\
      -\mathbf{ad}\left(\mbsbar{\gamma}_{ab}\right) \mbf{B} & \eye \ebm}_\textrm{$\mbf{L}$} \delta \mbf{w}_{ab}. \label{eq: lin_proc}
\end{align}
At this point, the linearized process model given by \eqref{eq: lin_proc}
 can be discretized using the matrix exponential \cite{barfoot2017state,farrell2008aided}. 
 The resulting discrete-time Jacobians are then used in a  
standard iterated EKF framework \cite{skoglund2015extended}.
\subsection{Measurement Model}
From \eqref{eq: point_corr}, given a point correspondence obtained from camera measurements
of a point feature $p_i$ lying on a plane, the pair can be related
by a homography transformation $\mbf{H}_k \triangleq \mbf{H}_{ab}$ via
\begin{equation}
    \mbf{y}_b^i = \mbf{g}(\mbf{H}_k\inv \mbf{p}_{a}^{i}) + \mbf{v}_k \label{eq: meas model},
\end{equation}
where $\mbf{p}_{a}^{i}$ represents the measurement of $p_i$ resolved in $\mathcal{F}_{a}$, 
in normalized image coordinates, $ \mbf{y}_b^i$ the noisy measurement in pixel 
coordinates of the same feature resolved 
in $\mathcal{F}_{b}$, and $\mbf{v}_k \sim \mathcal{N}\left(\mbf{0}, \mbf{R}_k\right) \in \rnums^2$ 
models white noise on the pixel measurement. Setting $\mbf{r}_{b}^{i} \triangleq
 \mbf{H}_k\inv \mbf{p}_{a}^{i}= \bbm x & y & z \ebm^\trans$,
\begin{equation} 
    \mbf{g}\left(\mbf{r}_{b}^{i} \right) \triangleq \f{1}{z}\mbf{D}\mbf{K}\mbf{r}_{b}^{i},
\end{equation}
where $\mbf{D} = \bbm \eye_{2\times2} & \mbf{0}_{2\times 1} \ebm$ and $\mbf{K}$ is the previously
introduced intrinsic parameter matrix to model a pinhole camera. 
The linearized measurement model is then
\begin{align}
    \delta\mbf{y}_b^{i} &\approx  \f{\partial  \mbf{g}(\mbf{r}_{b}^{i} )}
    {\partial \mbf{r}_{b}^{i} } \bbm\left.\frac{D \mbf{r}_{b}^{i}(\mbf{H}_k)}{ D \mbf{H}_k} \right|_{\mbfbar{H}_k}&\left.\frac{\partial \mbf{r}_{b}^{i}(\mbs{\Gamma}_k)}{ \partial \mbs{\Gamma}_k} \right|_{\mbsbar{\Gamma}_k} \ebm\delta \mbf{x}_k + \delta\mbf{v}_k,
\end{align}
where 
\begin{align}
    \f{\partial  \mbf{g}(\mbf{r}_{b}^{i} )}{\partial \mbf{r}_{b}^{i} } &= \f{1}{z}\bbm f_u & 0 & -f_u \f{x}{z} \\
        0 & f_v & -f_v \f{y}{z} \ebm, \\
        \bbm\left.\frac{D \mbf{r}_{b}^{i}(\mbf{H}_k)}{ D \mbf{H}_k} \right|_{\mbfbar{H}_k}&\left.\frac{\partial \mbf{r}_{b}^{i}(\mbs{\Gamma}_k)}{ \partial \mbs{\Gamma}_k} \right|_{\mbsbar{\Gamma}_k} \ebm &= \bbm {\mbfbar{H}_k\inv\mbf{p}_{a}^{i}}^\odot & \mbf{0}_{3 \times 8} \ebm .
\end{align}
With these Jacobians, 
the prediction and correction steps of the iterative EKF can be executed. 

\subsection{Robust Loss}
\label{sec: robustloss}

An iterated EKF's correction step can be formulated as a 
weighted nonlinear
least squares problem (WNLS) \cite{skoglund2015extended} 
solved using the Gauss--Newton algorithm. In particular, 
the error and cost function, respectively, associated with the WNLS problem are
\begin{align}
    \mbf{e}(\mbf{x}_k)^\trans &= \bbm \left(\mbfcheck{x}_k - \mbf{x}_k \right)^\trans &
    \left(\mbf{y}_k - \mbf{g}(\mbf{x}_k) \right)^\trans  \ebm, \\
    J(\mbf{x}_k) &= \f{1}{2} \mbf{e}(\mbf{x}_k)^\trans
                        \underbrace{\bbm \mbfcheck{P}_k\inv & \mbf{0}\\\mbf{0} & \mbf{R}_k\inv  \ebm}_\text{$\mbf{W}$}
                        \mbf{e}(\mbf{x}_k).
\end{align}
To mitigate the detrimental impact of outliers in the correction step, 
a robust loss is applied on the measurement residuals. 
It is straightforward to modify $\mbf{W}$ to add the robust loss weights,
as shown in \cite{mactavish2015all}. The \textit{SC/DCS} robust loss is chosen due to the
properties explained in \cite{mactavish2015all}. The \textit{SC/DCS} weight function is given by
$w(r) = \frac{4 c^2}{(c + r^2)^2}$ $\textrm{if } r^2\geq c$, otherwise, $w(r)=1$.

\section{Simulation Results}
\label{sec: sim}
\begin{figure*}[t!]
    \centering
    \includegraphics[width = 0.97\linewidth,trim={0.2em 2em 3em 1em},clip]{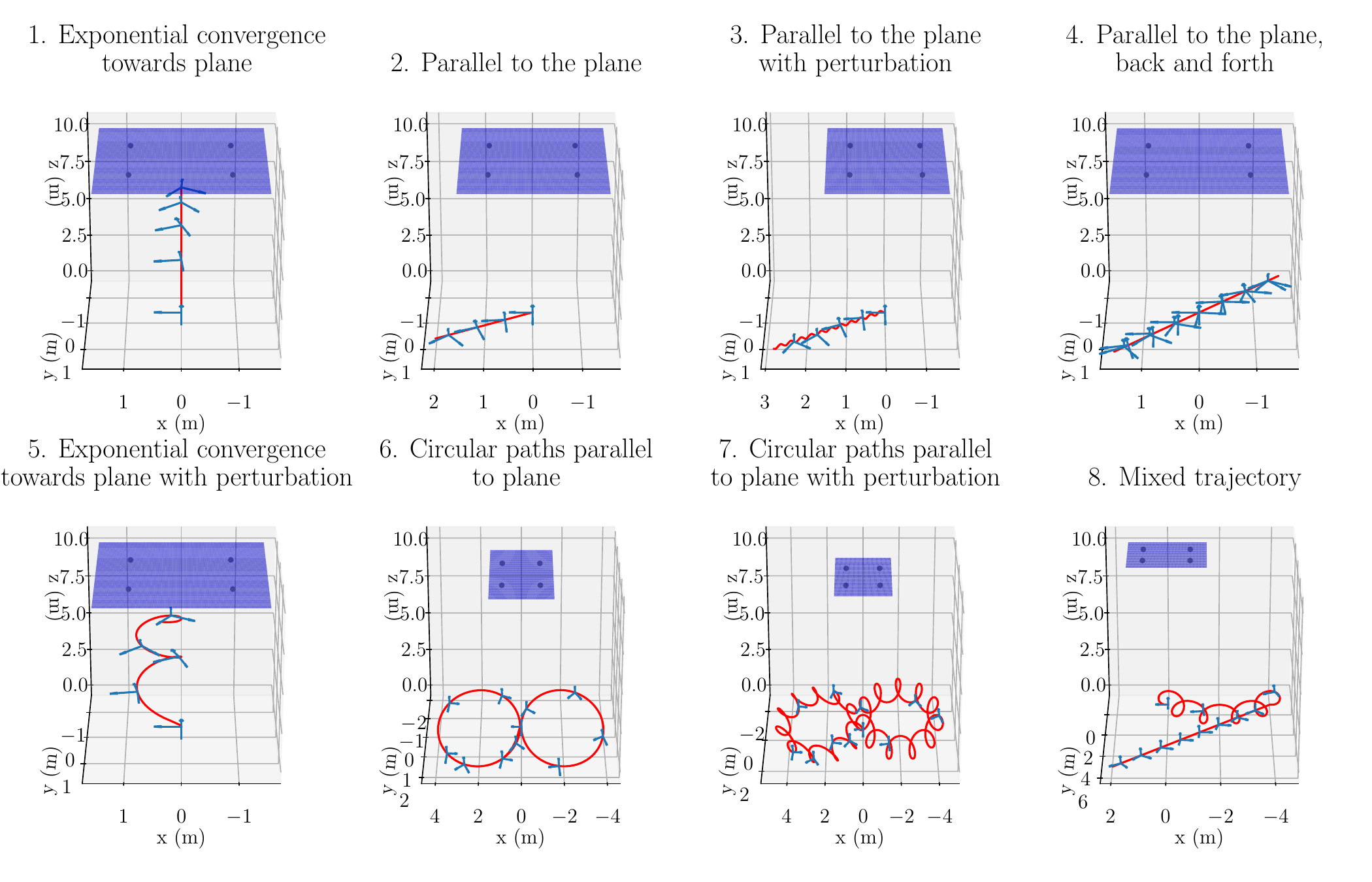}
    \caption{Simulated trajectories. Trajectories 1 and 2 do not violate
     the assumption that $\mbf{s}_a$ is constant. 
    The rest of the trajectories violate the assumption that $\mbf{s}_a$ is constant. }
    \label{fig: traj_sim}
 \end{figure*}
The proposed IMM filter is first tested in simulation. A set 
of trajectories
are generated, where a camera tracks 4 points on a plane at
 all times, the minimum
number of point matches to define a homography \cite{hartley2003multiple}.
The rate gyro and camera provide data at $\SI{90}{\Hz}$ and $\SI{30}{\Hz}$,
 respectively. Gyroscope and camera measurements are corrupted by additive Gaussian noise,
 simulated using $\mbc{Q}_g = \sigma_g^2 \eye$, $\mbf{R}_k = \sigma_r^2 \eye$, with
 $\sigma_g = \SI{0.01}{\radian \per \second}$,
 $\sigma_r = 1 \mathrm{\,pixel}$.
Initial uncertainty is set to $\mbfhat{P}_0 = \num{1e-1} \mbf{1}$.

100 Monte Carlo runs with varying
initial conditions and noise realizations are performed for each trajectory
to evaluate the
filters' consistency and accuracy when the assumption that $\mbf{s}_a$ is constant is broken 
in different ways. Among the tested filters are 2 versions of the iterated EKF.
The first is \textit{EKF tight}, which is confident in the
 assumption that $\mbf{s}_a$ is constant, by setting $\sigma_m^2 = \num{1e-7}$. 
 The second is \textit{EKF loose},
 with $\sigma_m^2 = \num{1e-1}$, which has little confidence in the assumption
  that $\mbf{s}_a$ is constant.
An IMM composed of both versions of the EKFs is also tested, as well as
 the observer from \cite{hua2019feature}.

\begin{figure}[thpb]
    \centering
    \includegraphics[width = \linewidth]{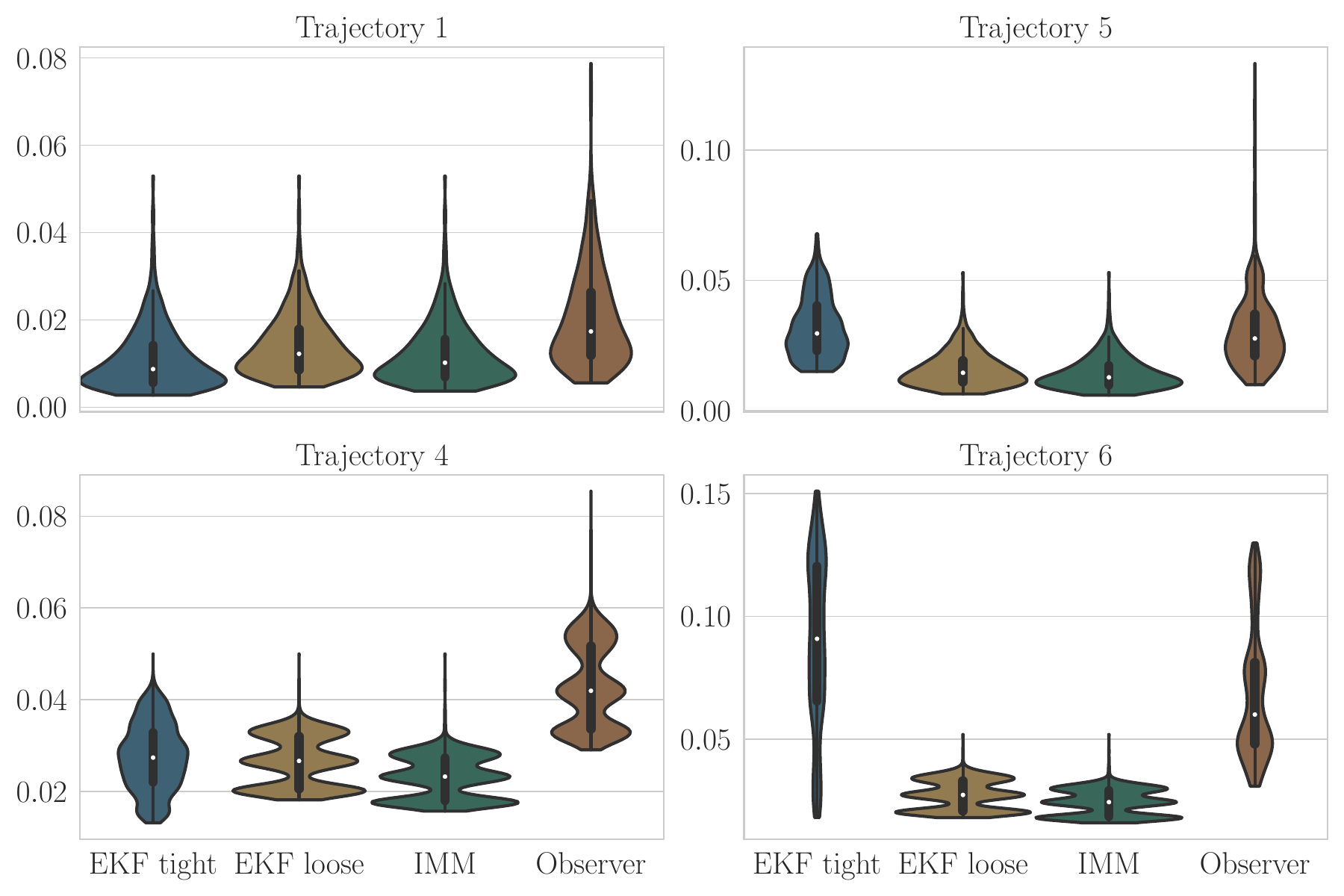}
    \caption{Violin plots to show distributions of $r_k$ in 4 sample trajectories from
    simulated data. EKF tight/loose, IMM, 
    and observer are tested. }
    \label{fig: violin}
 \end{figure}

 To evaluate the accuracy of the filters on each trajectory, the error
  \begin{equation}
    r_k =\|\log\left(
        \mbfhat{H}_k \mbf{H}_k\inv
    \right)^\vee\|_2 \label{eq: error_metric}
 \end{equation}
 is used to compare homography estimates to the true homography value at each time step.
 In Figure~\ref{fig: violin}, it is shown how $r_k$ is distributed across all time steps
in 4 trajectories from Figure~\ref{fig: traj_sim}.
\textcolor{black}{The trajectories displayed in Figure~\ref{fig: violin} are chosen to provide distinctive information about
the estimators' performance. For example, for Trajectories 1, 2, and 3, the plots look very similar,
so only Trajectory 1 out of 1, 2, and 3, is presented.} 
The average of $r_k$ across all trials and then averaged across all time steps
is displayed in Table~\ref*{tab: sim_results}. \textit{EKF tight} performs
the best on this metric when the trajectories respect the assumption
 that $\mbf{s}_a$ is constant as in 
Trajectories 1 and 2, or
closely do, as in Trajectory 3. In the remaining trajectories, the 
performance of \textit{EKF tight} degrades and becomes
the worst among the tested filters. In general, the IMM offers the best
performance when the assumption is less respected. Only in Trajectory 7,
 when the assumption is severely broken,
\textit{EKF loose} outperforms the IMM by a small margin. The observer of \cite{hua2019feature} 
has higher homography estimation error
in all trials, since all the filters are characterizing the sensor noise
properly and tuned accordingly. 
When tuning the observer it is observed 
that modifying the gains did not change the observer's performance drastically.
The observer is, in effect, constant gain, while the IMM filter changes
 the gain at each time step. 

\begin{figure}[t!]
    \centering
    \includegraphics[width = 0.95\linewidth]{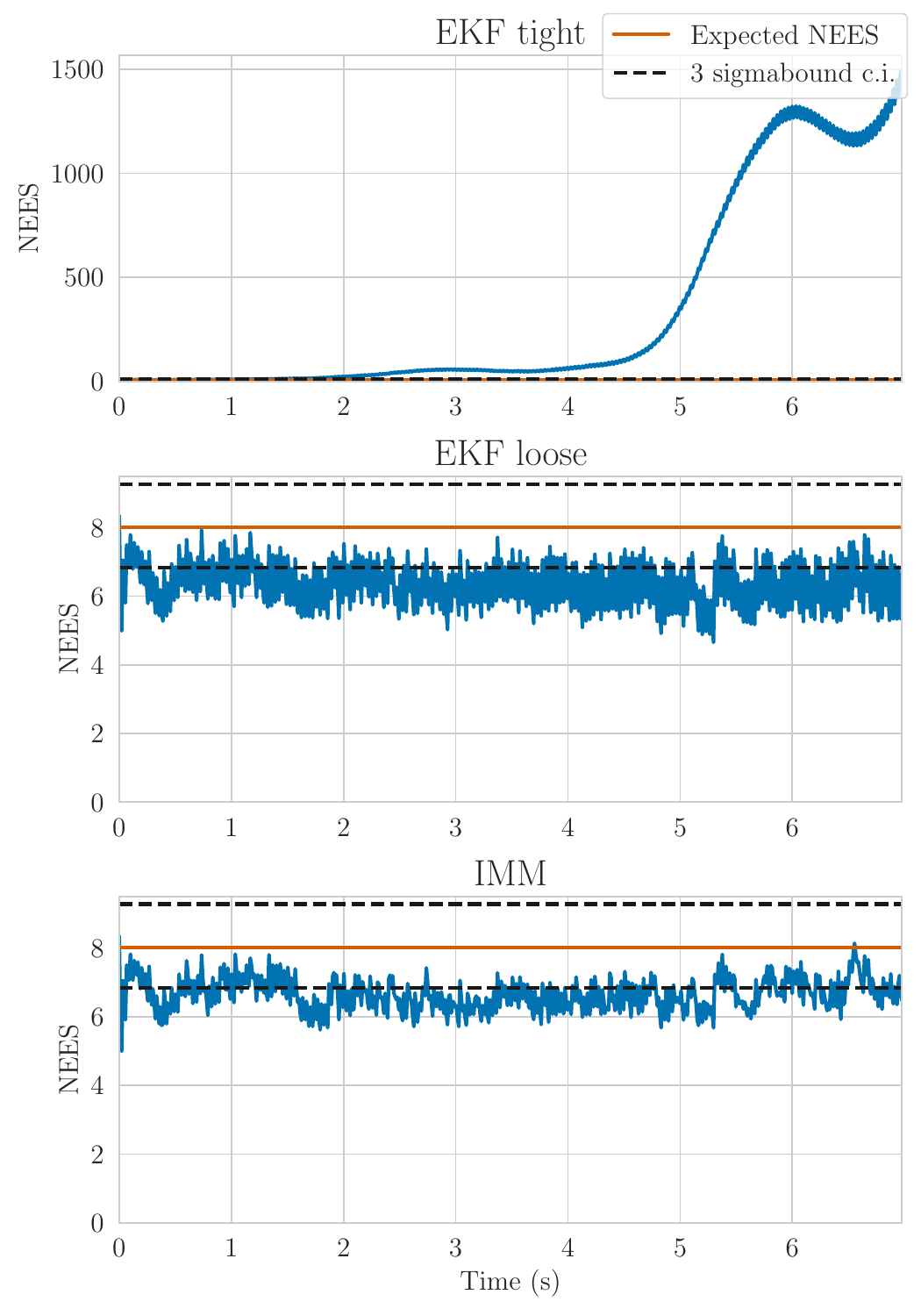}
    \caption{Example of NEES plots for Monte Carlo trials on Trajectory 5 
    comparing \textit{EKF tight}, \textit{EKF loose} and IMM approaches.}
    \label{fig: nees_imm}
 \end{figure}

Assessing consistency is done using the normalized
estimation error squared (NEES) test \cite[Sec. 5.4]{bar2001estimation} on Monte Carlo runs.  The
NEES test involves computing a $\epsilon_k \sim \chi^2_8$ statistic using the error
trajectory and the predicted covariance of such error, where
$\chi^2_8$ is the Chi-square distribution with 8 degrees of freedom,
\begin{align}
    \mbs{\xi}_k &= \log\left(
        \mbfhat{H}_k \mbf{H}_k\inv
    \right)^\vee, \qquad
    \epsilon_k = \mbs{\xi}_k^\trans \mbfhat{P}_{hh, k}\inv \mbs{\xi}_k,
\end{align}
where $\mbfhat{P}_{hh, k}$ is the block on the diagonal 
of $\mbfhat{P}_{k}$ corresponding to the homography state.
To assess with $3\sigma \approx 99.73 \%$ confidence if the estimator is consistent,
$\epsilon_k$ should remain between an upper and lower threshold \cite{bar2001estimation}.
In Figure~\ref{fig: nees_imm}, $\epsilon_k$ is averaged across all 100 trials,
and then plotted as a function of time. \textit{EKF tight} is a consistent estimator
 when the assumption that
$\mbf{s}_a$ is constant is respected, but as soon as the assumption is broken, the NEES values diverge,
 as shown in the top of Figure~\ref{fig: nees_imm}. This is
expected, since the errors are large. 
For \textit{EKF loose}, the NEES value is below the lower threshold, producing
inconsistent results since the covariance estimate is too large. 
However, having a large covariance is preferable
to having the error be too large for the covariance.
For the IMM, the NEES value goes below the lower threshold
 in some 
trajectories for the same reasons as
\textit{EKF loose} does, but stays closer to the lower threshold, indicating that the covariance
is better modulated than \textit{EKF loose},
\textcolor{black}{and providing a more consistent estimate. 
These results 
favour the implementation of the IMM over a single EKF
with a fixed process model noise level. Tuning the EKF is hard as the covariance is dependent on the level in which the motion breaks the assumption that $\mbf{s}_a$ is constant, and therefore an adaptive covariance is required.}

\begin{table}[t!]
    \caption{Performance of EKF tight/loose, IMM, and observer using simulated data. 
    The error $r_k$ is averaged across 100 Monte Carlo trials and then time steps.}
    \label{tab: sim_results}
    \begin{center}
    {\footnotesize
    \begin{tabular}{cccccc}
        \toprule
        & \multicolumn{4}{c}{$\Aver{r_k}$}  & \\
         \cmidrule(rl){2-5}
         \multicolumn{1}{p{1em}}{Traj.} &\multicolumn{1}{p{1em}}{\centering EKF \\tight}&
          \multicolumn{1}{p{1em}}{\centering EKF \\loose}  & IMM & Observer & \multicolumn{1}{p{1.5cm}}{\centering \% Diff. IMM/Obs.}\\
         \midrule
         1&\textbf{0.0108} & 0.0140 &  0.0121 & 0.0201  & 39.5\% \\
         2&\textbf{0.0212} & 0.0269 &  0.0237 & 0.0424 & 44.1\% \\
         3&\textbf{0.0207} & 0.0266  &  0.0235 & 0.0433 & 45.7\% \\
         4&0.0274 & 0.0266 &  \textbf{0.0231} & 0.0429 & 46.1\% \\
         5&0.0319 & 0.0158  &  \textbf{0.0143} & 0.0308 & 53.4\% \\
         6&0.0897 & 0.0273 &  \textbf{0.0242} & 0.0684 & 64.6\% \\
         7&0.1319 & \textbf{0.0383} &  0.0388 & 0.1268 & 69.4\% \\
         8 &0.0501 & 0.0299 & \textbf{0.0272} & 0.1062 & 74.4\% \\
        \bottomrule
        \end{tabular}
    }
    \end{center}
\end{table}



\section{Experimental Results}
\label{sec: exp}

\begin{figure*}[ht]
    \centering
    \begin{subfigure}[t]{2\columnwidth}
        \centering
    \includegraphics[width = 0.97\linewidth,trim={1em 1em 2em 0.5em},clip]{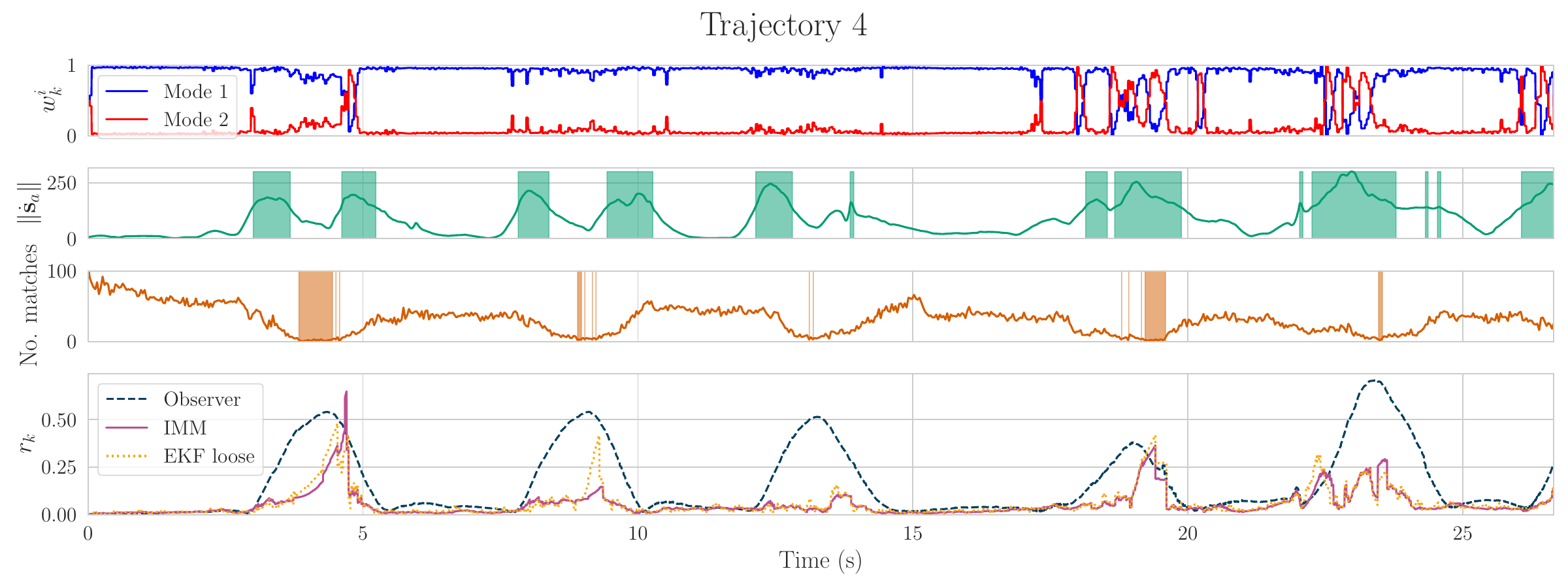}
    \label{fig: imm_trial3}
    \end{subfigure}
    \begin{subfigure}[t]{2\columnwidth}
        \centering
    \includegraphics[width =  0.97\linewidth, trim={1em 1em 2em 0.5em}, clip]{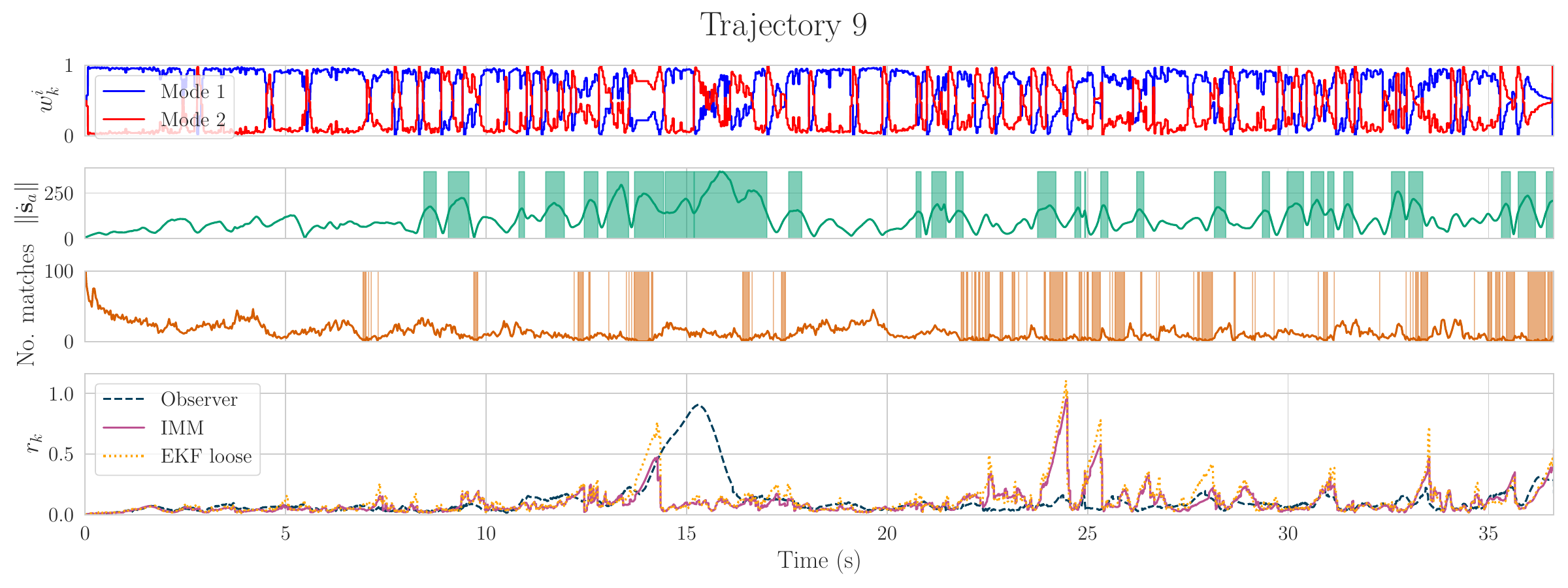}
    \label{fig: imm_trial1}
    \end{subfigure}
    \caption{IMM model probabilities depending on how the $\mbf{s}_a$ is constant assumption 
    is broken and number of 
    matches in \textcolor{black}{Trajectories 4 and 9}. Mode 1 corresponds to $\sigma_m^2 = 10^{-6}$ and mode 2 corresponds to $\sigma_m^2 = 1$.
     Evolution of IMM, \textcolor{black}{\textit{EKF loose}}, and observer's error $r_k$ is shown
    in bottom plots.  Green shaded regions represent
    time frames where $\|\mbfdot{s}_a \|>\alpha =155$.
    Orange shaded regions represent time frames where
     number of tracked features goes below 4. When $\|\mbfdot{s}_a\|> \alpha$, 
     the observer performance suffers. 
     When insufficient features are tracked, the EKF and IMM performances suffer. }
     \label{fig: imm}
 \end{figure*}

\begin{table}[thpb]
    \caption{\textcolor{black}{Performance of EKF loose, IMM, and observer using experimental data.
    The error $r_k$ is averaged across all time steps. }}
    \label{tab: exp_results}
    \begin{center}
        {\footnotesize
    \begin{tabular}{ccccc}
        \toprule
        & \multicolumn{3}{c}{$\Aver{r_k}$} &  \\
       \cmidrule(rl){2-4}
       \multicolumn{1}{p{1em}}{\centering Traj.} &
        \multicolumn{1}{p{1em}}{\centering EKF loose}  & IMM &
         Observer & \multicolumn{1}{p{1.5cm}}{\centering \% Diff. IMM/Obs.}\\
        \midrule
        
         1&0.0837 & \textbf{0.0381} & 0.0512 &      25.6 \%   \\
         2& \textcolor{gray}{\textbf{0.0679}} & \textcolor{gray}{\textbf{0.0691}} & 0.0923 &     25.1  \%  \\
         3&0.0681 & \textbf{0.0577} & 0.1352 &    57.3  \%   \\
         4&0.0595 & \textbf{0.0542} & 0.1619 &     66.5 \%  \\
         5&0.0910 & \textbf{0.0656} & 0.0787 &    16.7  \%  \\
         6&0.0582 & 0.0485 & \textbf{0.0434} &   -11.7  \%  \\
         7&0.0782 & \textbf{0.0607} & 0.0844 &    28.0  \%  \\
         8&0.0673 & 0.0498 & \textbf{0.0396} &   -25.6  \%  \\
         9&0.1176 & \textbf{0.1002} & 0.1145 &     12.6 \%  \\
        
         \bottomrule
        \end{tabular}
        }
    \end{center}
\end{table}

\begin{figure}[thpb]
    \centering
    \includegraphics[width = 0.95\linewidth]{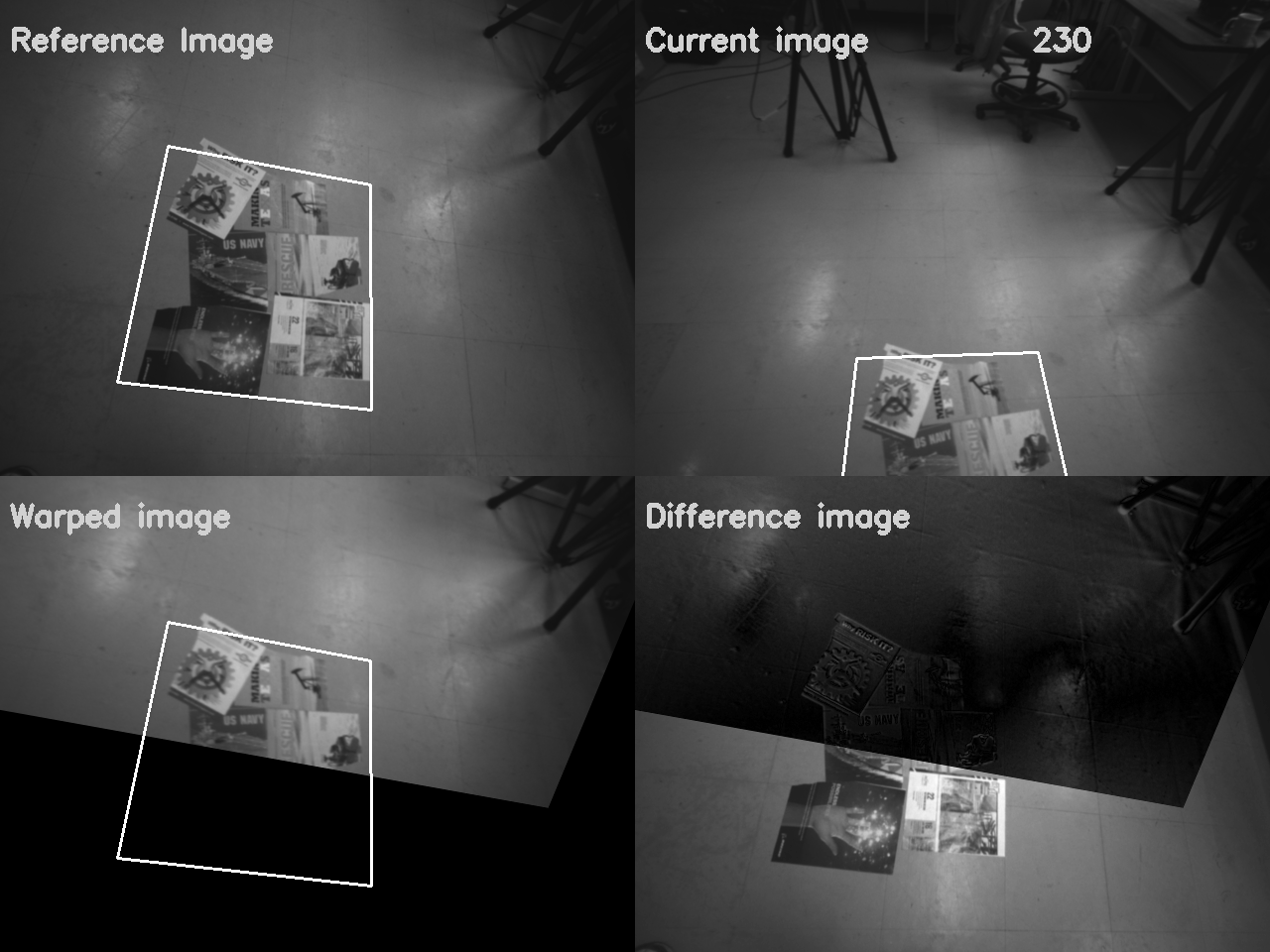}
    \caption{Experimental method that consists of tracking features on the ground. 
    The white square is used to visualize
    the estimated homography.
    The frame at the illustrated timestamp has the
      pattern, which is partially out of camera view, approximately
       realigned with the reference frame by the current homography matrix
        estimate.
    The current image is warped by the homography
     estimate on the bottom left.}
    \label{fig: trial_snap}
 \end{figure}


An 
Intel Realsense D435i is used to collect data. Angular 
velocity measurements are provided by a built-in rate gyro at 200 Hz.
The left camera of the stereo rig is used for camera measurements of $640\times 480$ pixels 
at 30 Hz. Ground truth data is collected using an OptiTrack optical motion
capture system at 120 Hz. The noise parameters used are
$\sigma_g = \SI{0.022}{\radian \per \second}$,
$\sigma_r = 1 \mathrm{\,pixel}$.

Nine trials are recorded, all about 1 minute long. 
In each trial the camera tracks a plane about $\SI{1.5}{\meter}$ away, consisting of magazine pages spread on the floor,
so salient features can be detected by the front-end system, as shown in Figure~\ref{fig: trial_snap}.
The camera moves above this plane while rotating, with different paces in all trials 
to test how well the filters work when the assumption that $\mbf{s}_{a}$ is
 constant in time
is broken. The camera does not
 observe the plane
for brief moments of time, and occlusions are also added in some trials
to assess the robustness of the proposed approaches.

\textcolor{black}{In order to obtain point correspondences for every tested estimator},
 ORB descriptors are used
in its \texttt{OpenCV} implementation \cite{rublee2011orb}. 
Since the goal is to estimate
homography \wrt a reference frame, the feature points from each image are
 all matched 
against those from a reference image, which is 
picked from the first few frames in the trials.
\textcolor{black}{RANSAC is used to remove outliers prior to 
the estimation process. The robust loss described in Section~\ref{sec: robustloss},
is also employed in all estimators to ensure any remaining outliers are indeed discarded.}

The state estimate is initialized as $\mbfcheck{H}_{0} = \eye, \mbscheck{\Gamma}_0 = \mbf{0}$ with 
$\mbfcheck{P}_0 = 10^{-4}\eye$ for all trials
since the reference frame was set as one of the frames recorded and the filter initialized at 
that same frame,
so the initial error is very small. The IMM is composed in this case of two iterated EKFs, 
the first one with $\sigma_m^2 = 10^{-6}$ and the second with $\sigma_m^2 = \num{1e-0}$. The 
robust loss function implemented to reject outliers 
is \textit{SC/DCS} from \cite{mactavish2015all}, with $c=9.5$. The transition probabilities are set
as $\mbs{\Pi} = \bbm 0.9 & 0.1 \\ 0.1 & 0.9\ebm$. \textcolor{black}{This value is picked heuristically
and is observed to not impact the IMM's performance drastically.}

The error $r_k$ defined in~\eqref{eq: error_metric}  is used
for performance evaluation. 
In Table~\ref{tab: exp_results}, it can be seen that in the recorded trajectories,
\textcolor{black}{the IMM has the best performance in seven out of nine trajectories, while the observer
is the best in the remaining two. The \textit{EKF loose} also outperforms the observer in 
four out of nine trajectories, and even having a similar performance to the IMM in one trajectory.
The \textit{EKF tight} is omitted
from the experimental results 
because in most of the experimental trials it diverged.}

 \textcolor{black}{To explain the 
varying performance across the recorded trajectories, Figure~\ref{fig: imm}
shows the performance of the estimators in two trajectories plagued with assumption breaks and occlusions. It can be observed how in Trajectory~4, in the time frames where the
assumption that
$\mbf{s}_a$ is constant 
is violated to a larger degree, the observer's error is bigger than the IMM, since the IMM can adapt its gain to handle the assumption violation. The 
\textit{EKF loose} has similar performance to the IMM in those time frames because it always has low confidence in the process model, but the performance is not as good when the assumption is respected, as shown in Section~\ref{sec: sim} and Table~\ref{tab: exp_results} results.}
\textcolor{black}{Trajectory~9 is challenging, with many occlusions and assumption violations. In the period going from $\SI{14}{\second}$ to $\SI{17}{\second}$
where the assumption is violated for an extended period of time, the observer performs poorly again. It appears from the rest of the trajectories, that when the assumption is broken for a significant period, the observer struggles.}

\textcolor{black}{During occlusions in both Trajectories 4 and 9, the estimators utilize the process model and the rate-gyro
measurements to propagate the homography estimate, and thus provide an estimate that otherwise would have been impossible to provide given only images. The \textit{EKF loose} has low confidence in the process model, causing the uncertainty on the estimate to grow significantly during an occlusion, and it can struggle to converge to a good estimate when enough measurements are available again. The IMM offers a compromise, by having an adaptive covariance, and not completely disregarding the process model. In Trajectory~4, around the $\SI{9}{\second}$ mark, it can be seen how the IMM puts more weight into Model 2, the same present in the \textit{EKF loose}, but not all the weight, thus ensuring a good performance. The IMM can in some cases put significant weight in Model 2, and produce behaviours similar to \textit{EKF loose}, but in general, the IMM performs better during occlusions than the \textit{EKF loose}, as can be seen during the many occlusions of Trajectory~9.
Concerning the observer, it usually had the best performance during occlusions, as shown in the second half of Trajectory~9. This can be, in part, explained by the theoretical guarantees of the observer \cite{hua2019feature}, which the EKF and IMM lack \cite{barrau2016invariant}.}



\textcolor{black}{Overall, the IMM offers the best compromise, by better handling occlusions and assumption violations than the \textit{EKF loose} and the observer, respectively. The observer shows good robustness to occlusions, but the IMM shows the best overall performance in varied trajectories while also offering covariance 
information and robustness during occlusions due to the utilization of a rate gyro, as shown in Figure~\ref{fig: imm}}.

\section{Conclusion}
\label{sec:conclusion}
In this paper, the problem of estimating homography using rate gyro and camera 
measurements is addressed. This paper's novelty lies
 in the use of the Bayesian filtering framework in concert with
  a simplified process model. 
In particular, two iterated EKFs are used within an IMM filter. 
The approach is compared to a nonlinear deterministic observer
 in both simulation and experiments where \textcolor{black}{an overall better}
  performance is realized. 
 The proposed Bayesian approach offers covariance information, 
 unlike the observer. 
A Bayesian approach opens the avenue 
for adaptive filtering, as in this letter, but also 
post-processing procedures
such as outlier removal, smoothing, loop-closure detection, 
and quality control.

\section*{Acknowledgment}

Thanks to M. A. Shalaby, C. Cossette, and
M. Cohen for their helpful feedback. Thanks to J. Trumpf, R. Mahony, and T. Hamel
for motivating homography estimation research. 




\bibliographystyle{IEEEtran}
\bibliography{refs}


\end{document}